%% file: template.tex
\documentclass[10pt,twoside]{article}

\usepackage[T1]{fontenc}
\usepackage[dvipsnames]{xcolor}
\usepackage[width=122mm,left=12mm,paperwidth=146mm,height=193mm,top=12mm,paperheight=217mm]{geometry}
\usepackage{amsmath}
\usepackage{amssymb}
\usepackage[numbers,sort&compress]{natbib}
\usepackage{xspace}
\usepackage[hyphens]{url}
\usepackage{etoolbox}
\usepackage[labelfont=bf,font=small,tableposition=bottom]{caption}
\usepackage[skip=3pt]{subcaption}
\definecolor{accvblue}{rgb}{0,0,1.0}
\usepackage{graphicx}
\usepackage{booktabs}
\usepackage{array}
\usepackage{makecell}
\usepackage{standalone}
\usepackage[accsupp]{axessibility}
\usepackage[pagebackref,breaklinks,colorlinks,citecolor=accvblue]{hyperref}
\usepackage[capitalize]{cleveref}

\makeatletter
\DeclareRobustCommand\onedot{\futurelet\@let@token\@onedot}
\def\@onedot{\ifx\@let@token.\else.\null\fi\xspace}
\makeatother

\newcommand{\keywords}[1]{\par\smallskip\noindent\textbf{Keywords: }{\def\and{\unskip; }#1}}

\date{}

\newcommand{\authorblock}[3]{\begin{minipage}[t]{#1}
    \centering
    \textbf{#2}\\[0.2em]
    #3
  \end{minipage}}

\makeatletter
\renewcommand{\maketitle}{\par
  \noindent\rule{\textwidth}{2.5pt}\par
  \vspace{1.2em}
  {\centering
    \Large\bfseries \@title\par
  }
  \vspace{1.2em}
  \noindent\rule{\textwidth}{0.5pt}\par
  \vspace{2.2em}
  \noindent
  \makebox[\textwidth][c]{\@author}\par
  \vspace{1.5em}
}
\makeatother

\newcolumntype{C}{>{\centering\arraybackslash}p{1.75cm}}

\newcommand{\best}[1]{\textbf{#1}}
\newcommand{\meanstd}[2]{#1\,\raisebox{-0.35ex}{\scalebox{0.5}{$\pm\,#2$}}}

\title{Stable Unsupervised Continual Chunking with Sheaf SyncMap}
\author{
  \authorblock{0.4\textwidth}
    {Xueyuan Li}{Kyushu University}
  \authorblock{0.4\textwidth}
    {Danilo Vasconcellos Vargas}{Kyushu University}
}
\begin{document}

\maketitle

\begin{abstract}
Unsupervised Continual chunking is a fundamental problem in machine learning and neuroscience, where the goal is to identify groups of states that frequently co-occur in temporal sequences. A key challenge is to form accurate chunks while maintaining their stability over time. In this work, we propose sheaf regularization to reduce local inconsistencies in Decentralized SyncMap, a self-organizing system, and thereby stabilize its chunking dynamics. We introduce a radial sheaf structure that penalizes distance-dependent radial motion between pairs of variables. With two-state memory, the proposed method performs comparably to Decentralized SyncMap, with mean normalized mutual information (NMI) scores. Under dynamic memory, it achieves the highest NMI among the evaluated SyncMap variants on 17 of 18 probabilistic Continual General Chunking Problem (CGCP) graphs. The largest gains occur under dynamic memory and on graphs containing many small chunks. In the sequential adaptation experiment, Sheaf SyncMap also achieves high NMI after shifts in the input distribution, indicating that it can adapt to new knowledge while avoiding the negative transfer commonly observed in modern machine learning systems such as neural networks.
\end{abstract}

\noindent\textbf{CCS CONCEPTS}
\textbullet\ Machine learning;
\textbullet\ Bio-inspired approaches;
\textbullet\ Unsupervised learning


\section{Introduction}
Self-organization is a fascinating phenomenon widely observed in nature, in which complex structures emerge from simple interactions among individual components \cite{Bonabeau1997, Karsenti2008, Khadka2018}. Unlike many conventional engineered structures, which remain largely fixed after construction, self-organized biological structures can continuously reorganize in response to environmental changes. Examples include living bridges formed by army ants, mechanically adaptive clusters formed by honeybees, and transport networks formed by slime mold \cite{Reid2015, Peleg2018, Tero2010}. This ability to maintain functionality through decentralized adaptation motivates us to pursue self-organization as a principle for building robust and flexible artificial systems. A fundamental challenge, however, is how to preserve previously formed global structures when individual components have access only to limited local observations that do not explicitly encode those structures.

SyncMap is a self-organizing system that recognizes co-occurring patterns in temporal sequences and encodes them as spatially chunked representations in an unsupervised manner \cite{Vargas2021}. It can reorganize itself based on current observations, mimicking the adaptive behavior of biological systems. Standard SyncMap divides all variables into positive and negative groups, whose members are attracted to and repelled from their respective group centers. However, the standard SyncMap tends to collapse variables toward their corresponding centers, thereby losing local information within each chunk. Decentralized SyncMap replaces the two group centers with pairwise forces, preserving more local structure than center-based updates \cite{Li2025}. Despite this advantage, the limited observation window causes unobserved variables that have already formed a chunk to experience substantial repulsion, causing the chunk to spread and eventually vanish.

Recently, sheaf theory has attracted increasing attention in the machine learning community \cite{Barbero2022, Bodnar2022, Seely2026}. It provides a mathematical framework for resolving global inconsistencies by reducing local conflicts. Specifically, it assigns vector spaces to vertices and edges, defines restriction maps between them, and iteratively updates the vertex data to reduce local inconsistencies \cite{Zariski1956, Shepard1985}. In this work, we propose a sheaf regularization method to stabilize the chunking dynamics of Decentralized SyncMap. We introduce a radial sheaf structure that penalizes radial motion between pairs of variables based on their distance and repeller status. Experimental results show that the chunk structure in Decentralized SyncMap with sheaf regularization vanishes significantly more slowly. We also propose a directional history repulsion mechanism that stabilizes the sequential chunking dynamics by distinguishing interactions between positive and negative variables from those between pairs of positive variables and pairs of negative variables.

\begin{figure}[t]
  \centering
  \includegraphics[width=0.90\textwidth]{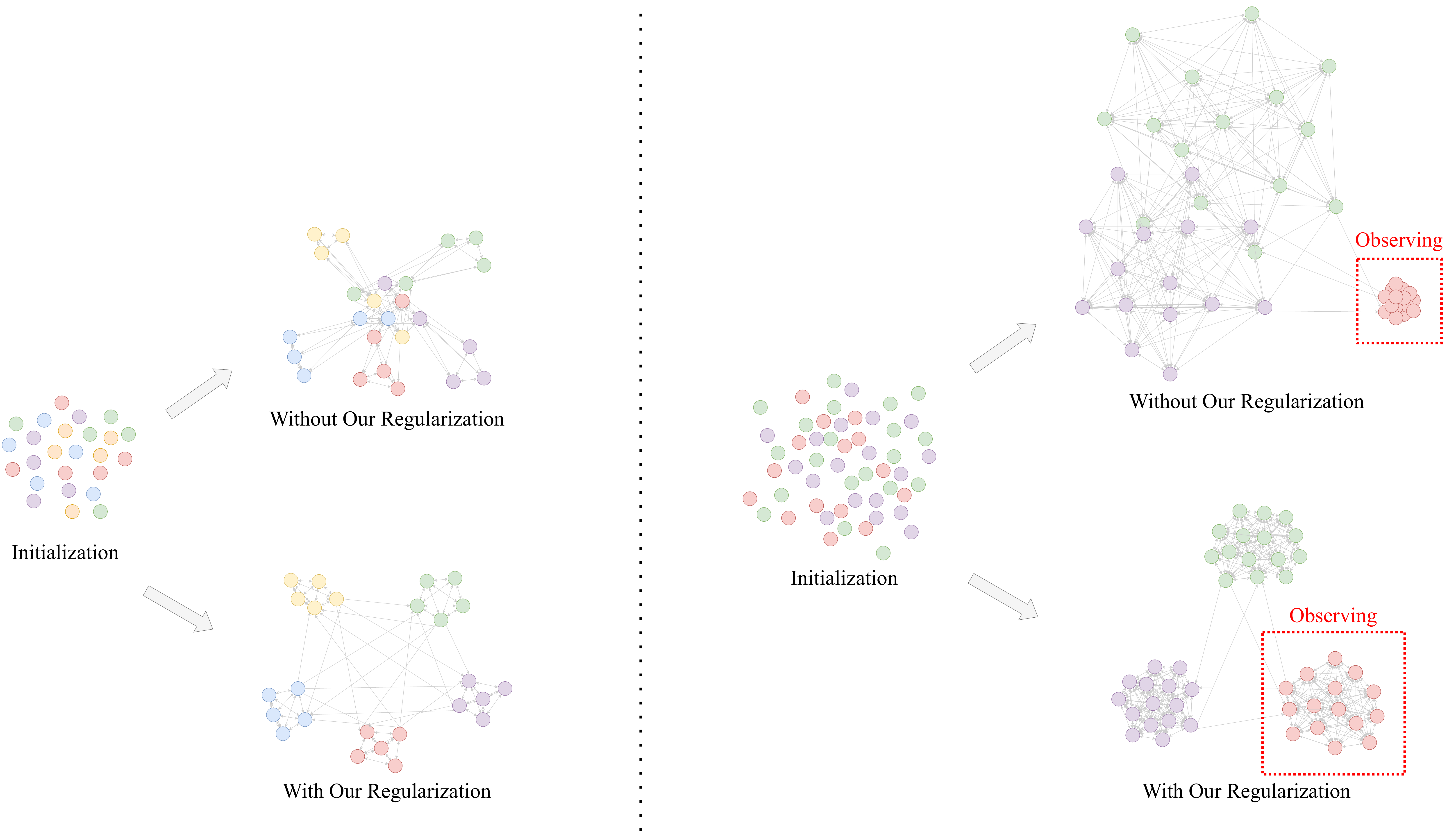}
  \caption{Overview of Sheaf SyncMap. Left panel: The number of chunks is greater than the number of variables per chunk. Without our regularization, variables that interact with other chunks become entangled, preventing the formation of correct chunks. Right panel: The number of chunks is smaller than the number of variables per chunk. Without our regularization, the variables being observed collapse to a single point, while the other chunks expand and eventually vanish. Sheaf regularization stabilizes the chunking dynamics in both cases.}
  \label{fig:abstract}
\end{figure}

\section{Related Work}
\subsection{Decentralized SyncMap}

The original formulation, referred to here as standard SyncMap, updates the position of each variable based on its interaction with the positive and negative centers:
\begin{equation}
  x_{i,t+1} = x_{i,t} + \alpha\left(
  \frac{\phi_{i,t}(cp_t-x_{i,t})}{|cp_t-x_{i,t}|}
  -
  \frac{(1-\phi_{i,t})(cn_t-x_{i,t})}{|cn_t-x_{i,t}|}
  \right),
  \label{eq:original_syncmap}
\end{equation}
where $x_{i,t}$ is the position of variable $i$ at time $t$, $\alpha$ is the adaptation rate, and $cp_t$ and $cn_t$ are the positive and negative centers at time $t$, respectively:
\begin{equation}
  cp_t = \frac{\sum_{i \in PS_t} x_{i,t}}{\left|PS_t\right|},
  \qquad
  cn_t = \frac{\sum_{i \in NS_t} x_{i,t}}{\left|NS_t\right|},
  \label{eq:centers}
\end{equation}
where $PS_t$ and $NS_t$ are the sets of variables in the positive and negative groups at time $t$, respectively. Variables in the positive group are called attractors, whereas variables in the negative group are called repellers. The indicator function $\phi_{i,t}$ is defined as:
\begin{equation}
  \phi_{i,t} =
  \begin{cases}
    1, & i \in PS_t, \\
    0, & i \in NS_t.
  \end{cases}
  \label{eq:indicator_function}
\end{equation}
During the learning process, the attractors are attracted toward the positive center, whereas the repellers are repelled from the negative center. One of the main weaknesses of the original SyncMap is that variables tend to move toward their corresponding centers and eventually collapse to nearly the same position, thereby losing local information within each chunk \cite{Vargas2021}. We also use Symmetrical SyncMap as a second center-based baseline, which has better performance than Standard SyncMap on imbalanced CGCPs \cite{Zhang2023}.

Decentralized SyncMap replaces the two group centers with pairwise forces \cite{Li2025}. Let $N$ represent the number of variables, and let $\mathbf{v}_t \in \{0,1\}^{N}$ denote whether each variable currently belongs to the positive or negative group (1 for positive, 0 for negative). We refer to $\mathbf{v}_t$ as the activation state. Therefore, $\bar{\mathbf{v}}_t=\mathbf{1}-\mathbf{v}_t$, and
\begin{equation}
  \mathbf{P}_t=\mathbf{v}_t\mathbf{v}_t^{\top}\in \mathbb{R}^{N\times N},\qquad\mathbf{M}_t=\bar{\mathbf{v}}_t\bar{\mathbf{v}}_t^{\top}\in \mathbb{R}^{N\times N},\qquad d_{ij,t}=\|\mathbf{x}_{i,t}-\mathbf{x}_{j,t}\|_2\in \mathbb{R}.
  \label{eq:decentralized_masks}
\end{equation}
Here, $P_{ij,t}$ selects attractor--attractor pairs at time $t$, whereas $M_{ij,t}$ selects repeller--repeller pairs. Their attraction and repulsion magnitudes are
\begin{equation}
  a_{ij,t}=C_{ij,t}\left(1+\gamma_{+}e^{-d_{ij,t}/r_{+}}\right),\qquad r_{ij,t}=M_{ij,t}\gamma_{-}e^{-d_{ij,t}/r_{-}}+\mu_H H_{ij,t},
  \label{eq:decentralized_pair_weights}
\end{equation}
where $C_{ij,t}\geq 0$ combines the current interactions between attractors with recency-weighted attractor history. The first term of $r_{ij,t}$ represents instantaneous repulsion from repellers, whereas $H_{ij,t}$ stores persistent repeller history weighted by a factor $\mu_H$.

For the original symmetric history rule, this memory evolves elementwise as
\begin{align}
  \widetilde{\mathbf{H}}_t&=\operatorname{clip}_{[0,1]}\left(\mathbf{H}_{t-1}+T_H^{-1}\mathbf{M}_t\right),
  \label{eq:decentralized_history}\\
  \mathbf{H}_t&=\left(\mathbf{1}\mathbf{1}^{\top}-\mathbf{P}_t\right)\odot\widetilde{\mathbf{H}}_t.
  \label{eq:history_clear}
\end{align}
A repeller pair therefore accumulates $1/T_H$ for each co-repelling step, up to a maximum of one, and its memory is cleared when both endpoints become attractors. A stored attractor--repeller entry remains active because it is neither accumulated by $\mathbf{M}_t$ nor cleared by $\mathbf{P}_t$.

The coordinate update is given by the sum of all pairwise forces:
\begin{equation}
  \mathbf{u}_{i,t}=\sum_{j\ne i}\frac{\mathbf{x}_{i,t}-\mathbf{x}_{j,t}}{d_{ij,t}}\left(\beta r_{ij,t}-\alpha a_{ij,t}\right),\qquad\mathbf{x}_{i,t+1}=\mathbf{x}_{i,t}+\eta\mathbf{u}_{i,t},
  \label{eq:decentralized_update}
\end{equation}
followed by global coordinate normalization. Although pairwise updates preserve local structure better than center-based updates, unconstrained relative motion can still cause chunking instabilities such as expansion, contraction, and breathing. This limitation motivates the sheaf regularization introduced below.

\subsection{Cellular Sheaf}

A sheaf is a mathematical framework that provides a way to resolve global inconsistencies by reducing local conflicts \cite{Zariski1956}. A cellular sheaf extends graph theory by applying sheaf theory to graphs, thereby enabling more complex interactions between vertices and edges \cite{Shepard1985}. A standard graph $G=(V,E)$, where $V$ is the set of vertices and $E$ is the set of edges, only specifies whether $u,v\in V$ are connected by an edge $e\in E$. By introducing a cellular sheaf $\mathcal{F}$, we can assign a vector space $\mathcal{F}(v)$ to each vertex $v$ and a vector space $\mathcal{F}(e)$ to each edge $e$. These vector spaces are called stalks. If data samples $x_u\in\mathcal{F}(u)$ and $x_v\in\mathcal{F}(v)$ are assigned to two neighboring vertices $u$ and $v$ connected by an edge $e$, their relationship is described by the restriction maps $\mathcal{F}_{u\trianglelefteq e}$ and $\mathcal{F}_{v\trianglelefteq e}$:
\begin{align}
  \mathcal{F}_{v\trianglelefteq e}:\mathcal{F}(v)\to\mathcal{F}(e), \quad \mathcal{F}_{u\trianglelefteq e}:\mathcal{F}(u)\to\mathcal{F}(e).\label{eq:sheaf_map}
\end{align}
These maps specify how information is mapped from vertices to edges. If $x_u$ and $x_v$ satisfy the consistency condition
\begin{equation}
  \mathcal{F}_{v\trianglelefteq e}(x_v)=\mathcal{F}_{u\trianglelefteq e}(x_u),
  \label{eq:section}
\end{equation}
then $(x_u,x_v)$ forms a section over the edge. If a collection of vectors satisfies the consistency condition for all edges in the graph, it forms a global section of the sheaf.
The strain $s_e$ is defined as follows to quantify local inconsistency:
\begin{equation}
  s_e=(\delta x)_e=\mathcal{F}_{v\trianglelefteq e}(x_v)-\mathcal{F}_{u\trianglelefteq e}(x_u),
  \label{eq:sheaf_strain}
\end{equation}
where $\delta$ is the coboundary operator of the sheaf. We use implicit sheaf-Laplacian diffusion to update the data associated with each vertex:
\begin{equation}
  x_{k+1}=(I+\eta L)^{-1}x_k,
  \qquad
  L=\delta^\top\delta,
  \label{eq:sheaf_diffusion}
\end{equation}
thereby reducing local inconsistencies and promoting global consistency.

\subsection{Continual General Chunking Problem}
The Continual General Chunking Problem (CGCP) was first proposed in \cite{Vargas2021}. It provides a unified formulation for several problems studied in neuroscience and computer science, including chunking, causal and temporal community detection, and unsupervised feature learning from temporal sequences.
CGCP aims to identify groups of states that frequently co-occur in temporal sequences while allowing the underlying data-generation process, or data structure, to change over time. The input sequence is generated by a random walk on a graph, where transitions between states follow a first-order Markov chain.

For our experiments, we synthesize 18 probabilistic CGCP graphs, each corresponding to a factor pair satisfying $KM=600$, with $3\leq K\leq120$ chunks and $5\leq M\leq200$ variables per chunk; thus, every graph has $N=600$ variables. The graph name \texttt{probabilistic$K$\_$M$} records these values; for example, \texttt{probabilistic120\_5} contains 120 chunks of five variables. Each chunk forms a complete subgraph without self-loops, and its last variable links to the first variable of every chunk, keeping the full graph connected. Trajectories are random walks over outgoing neighbors. Each visited state is represented by a one-hot vector, and state memory forms an activation vector from the Boolean union of a recent window. We use either $m=2$ states (current and previous) or $m_{\mathrm{dyn}}=\lfloor\min(30,\max(2,0.1N))\rfloor=30$. 
\subsection{Mutual Information}

We use Normalized Mutual Information (NMI) to evaluate the quality of chunking results \cite{Strehl2002}. Consider two random variables $X$ and $Y$. Their mutual information $I(X;Y)$ is defined as:
\begin{equation}
  I(X;Y)=\sum_{x,y}p(x,y)\log\frac{p(x,y)}{p(x)p(y)}.
  \label{eq:mutual_information}
\end{equation}
To normalize the mutual information with respect to the entropies of $X$ and $Y$, NMI is defined as:
\begin{equation}
  \text{NMI}(X;Y)=\frac{2I(X;Y)}{H(X)+H(Y)}.
  \label{eq:normalized_mutual_information}
\end{equation}
The NMI value ranges from 0 to 1, with a larger value indicating greater similarity between the two clusterings. NMI can therefore be used to evaluate the similarity between a predicted clustering and the ground-truth clustering. It is invariant to permutations of cluster labels, meaning that two identical clusterings receive the same NMI score even if their cluster labels are different.
We obtain the predicted clustering by applying DBSCAN to each recorded embedding \cite{Schubert2017}. DBSCAN uses a neighborhood radius $\epsilon$ and a minimum neighborhood size, \texttt{min\_samples}. Selecting $\epsilon$ by maximizing NMI against the ground-truth labels gives an oracle density-scale evaluation because the labels determine the clustering scale.

\section{Methodology}
\subsection{Directional history repulsion}

The pairwise temporal memory of repellers proposed in \cite{Li2025} is defined in \cref{eq:decentralized_history}. The term $\left(\mathbf{1}\mathbf{1}^{\top}-\mathbf{P}_t\right)$ in \cref{eq:history_clear} clears the bidirectional memory of all attractor--attractor pairs. To incorporate sequential information, we propose clearing only the directional memory associated with the currently activated attractor.
For instance, suppose the observed transition is from variable $i$ to variable $j$. In this case, only the memory entry $H_{ij,t}$ is set to zero, while $H_{ji,t}$ is preserved. Therefore, \cref{eq:history_clear} is modified as follows:
\begin{equation}
  \mathbf{H}_t=\left(\mathbf{1}\mathbf{1}^{\top}-e_i e_j^{\top}\right)\odot\widetilde{\mathbf{H}}_t.
  \label{eq:directional_history_clear}
\end{equation}
Here, $e_i$ and $e_j$ are the one-hot vectors corresponding to variables $i$ and $j$, respectively.
However, the pairwise history-repulsion term used in \cref{eq:decentralized_pair_weights} must remain symmetric. Therefore, we re-symmetrize the history-repulsion matrix before computing the pairwise forces:
\begin{equation}
  \mathbf{H}^{\text{eff}}_t=\frac{\mathbf{H}_t+\mathbf{H}_t^\top}{2}.
  \label{eq:directional_history_re_symmetrize}
\end{equation}

For the proposed directional-history variant, $H^{\mathrm{eff}}_{ij,t}$ replaces $H_{ij,t}$ in \cref{eq:decentralized_pair_weights}; hence,
\begin{equation}
  r_{ij,t}
  = M_{ij,t}\gamma_- e^{-d_{ij,t}/r_-}
    + \mu_H H^{\mathrm{eff}}_{ij,t}.
\end{equation}
The asymmetric matrix $\mathbf{H}_t$ remains the persistent history state used in subsequent updates. Symmetrization is applied only when computing the pairwise forces and does not overwrite the directional history.

\subsection{Radial Sheaf SyncMap}

\begin{figure}[h!]
  \centering
  \includegraphics[width=0.40\textwidth]{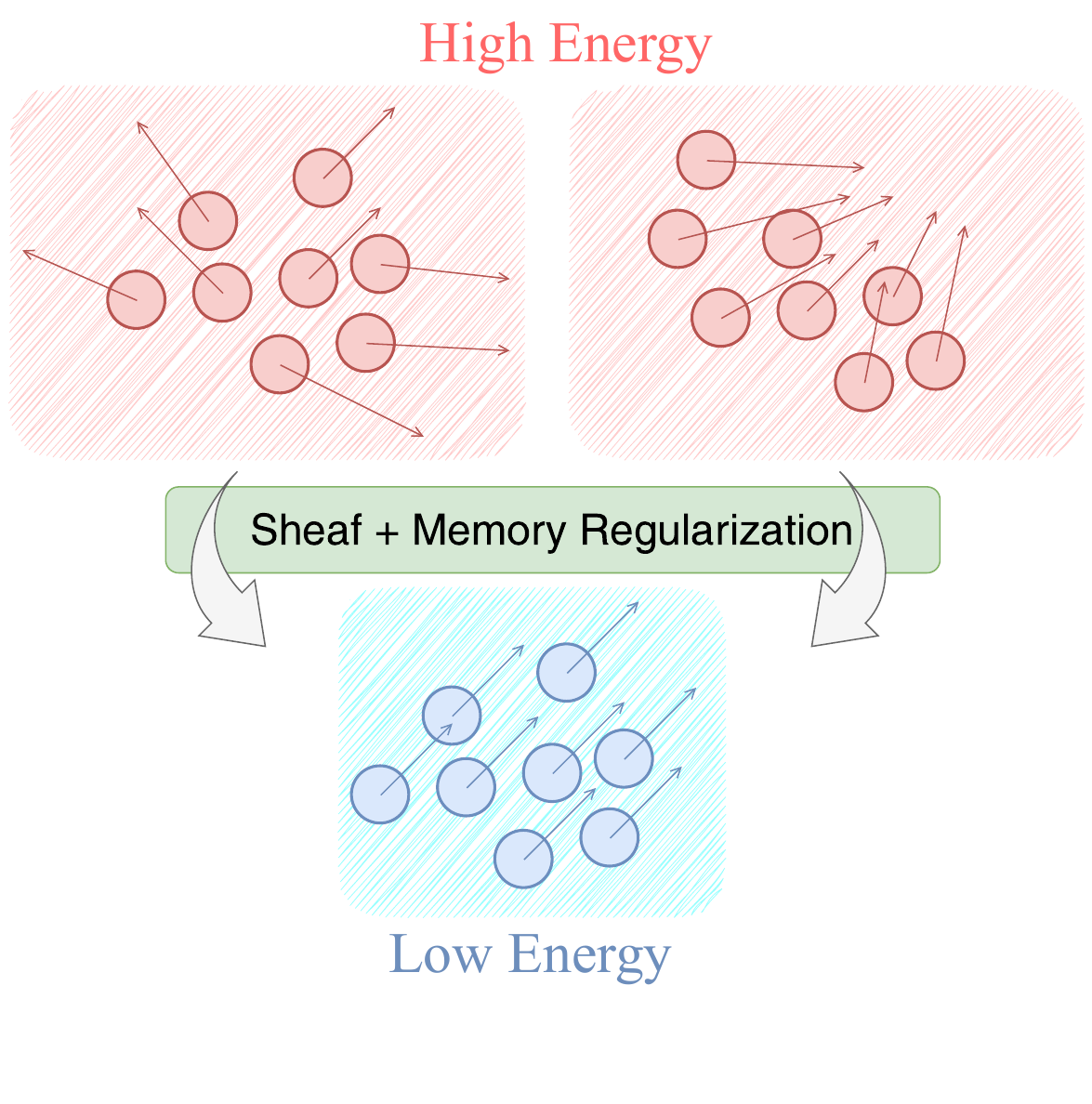}
  \caption{Intuition of the radial sheaf structure, where the arrows represent the directions in which the variables are moving. It suppresses the radial velocity between variables while leaving their velocities unchanged when they move together in the same radial direction.}
  \label{fig:sheaf_intution}
\end{figure}

In this work, we propose a radial sheaf structure to regularize the pairwise forces in Decentralized SyncMap. It acts as a regularizer to prevent chunks from expanding and eventually disappearing by suppressing pairwise radial velocities. Intuitively, it reduces the overall pairwise radial velocity while preserving the original velocity as much as possible (\cref{fig:sheaf_intution}). Therefore, it can help resist chunk expansion or collapse.
Each variable is associated with a node of the sheaf, and the edge set $E$ contains all unordered pairs of variables. The regularizer therefore applies relative radial geometry to every pair. The stalk of node $i$ is defined as
\begin{equation}
  \mathcal{F}(i)=\mathbb{R}^{\text{d\_model}},
  \label{eq:radial_sheaf_stalk}
\end{equation}
where $\text{d\_model}$ is the dimensionality of the SyncMap embedding space. We use $\text{d\_model}=3$ for all evaluated methods to provide a controlled comparison in a common low-dimensional space and to permit direct visualization of the embedding dynamics. Each node stalk stores the velocity of the corresponding variable. The edge stalk represents the relative radial velocity between two variables. Therefore, the strain $s_e$ associated with edge $e=(i,j)$ is naturally defined as
\begin{equation}
  q_e=\frac{x_i-x_j}{\lVert x_i-x_j\rVert^2}, \qquad s_e(v)=q_e^\top(v_i-v_j),
  \label{eq:radial_sheaf_edge_stalk}
\end{equation}
where $q_e$ is the distance-weighted radial direction between the two nodes, and $v_i$ denotes the candidate corrected velocity of node $i$. 
Writing $d_{ij}=\lVert x_i-x_j\rVert$ and $u_e=(x_i-x_j)/d_{ij}$, we have $q_e=u_e/d_{ij}$. In particular, when $v_i=\dot{x}_i$ and $d_{ij}>0$, $s_e(v)=\dot{d}_{ij}/d_{ij}=\frac{\mathrm{d}}{\mathrm{d}t}\log d_{ij}$.

The sheaf-Laplacian regularization is designed to reduce radial strain while keeping the modified velocity close to the original proposed velocity. Mathematically, the objective function $\mathcal{J}(v)$ is defined as
\begin{align}
  s_i &=
  \begin{cases}
    S, & i\text{ is a repeller},\\
    1, & i\text{ is not a repeller},
  \end{cases}
  \label{eq:repeller_scale}\\
  \omega_e &=\frac{s_i+s_j}{2}\cdot b^{1-\lVert x_i-x_j\rVert},
  \label{eq:edge_weight}\\
  \mathcal{J}(v)&=\frac12\lVert v-p\rVert_F^2 +\frac{\lambda}{2}\sum_{e\in E}\omega_e s_e(v)^2.
  \label{eq:radial_sheaf_objective}
\end{align}
The repeller scaling factor $s_i$ is introduced to regularize repellers more strongly than attractors. When a repeller interacts with an attractor, the term $(s_i+s_j)/2$ in \cref{eq:edge_weight} produces an intermediate weight between those of attractor--attractor and repeller--repeller pairs. The first term in \cref{eq:radial_sheaf_objective} keeps the modified velocity $v$ close to the original proposed velocity $p$, whereas the second term penalizes radial inconsistency. The hyperparameter $\lambda$ controls the trade-off between these two terms.
We then define the sheaf coboundary operator
$B\in\mathbb{R}^{|E|\times N\text{d\_model}}$. For an edge $e=(i,j)$,
\begin{equation}
  (Bv)_e=q_e^\top v_i-q_e^\top v_j=q_e^\top(v_i-v_j)=s_e(v),
  \label{eq:boundary_operator}
\end{equation}
where
$v=(v_1,v_2,\ldots,v_N)^\top
\in\mathbb{R}^{N\text{d\_model}}$
contains the velocities of all nodes. Let $W$ denote the diagonal matrix of edge weights $\omega_e$. The distance normalization in $q_e$ and the edge weight $\omega_e$
serve distinct roles. The former measures relative distance change,
whereas the latter controls the importance of each edge through
repeller scaling and distance decay. Minimizing \cref{eq:radial_sheaf_objective} yields
\begin{equation}
  \left(I+\lambda B^\top W B\right)\widetilde{v}=p,
  \label{eq:radial_sheaf_solution}
\end{equation}
where $B^\top W B$ is the radial sheaf Laplacian $L_{\mathcal F}$. Since $\lambda L_{\mathcal F}$ is positive semidefinite, $I+\lambda L_{\mathcal F}$ is positive definite, and therefore the solution $\widetilde{v}$ is unique. In this work, we solve \cref{eq:radial_sheaf_solution} using the conjugate gradient (CG) method. We initialize the solution with $\widetilde{v}_0=p$ and compute the residual at iteration $t$ as
\begin{equation}
  r_t
  =
  p-(I+\lambda L_{\mathcal F})\widetilde{v}_t.
  \label{eq:radial_sheaf_residual}
\end{equation}
CG iteratively constructs conjugate search directions from the residuals and updates $\widetilde{v}_t$ using an analytically determined step size. This procedure produces a corrected velocity $\widetilde{v}$ that remains close to the original proposed velocity $p$ while reducing radial strain, thereby improving the stability of the chunking dynamics.

\subsection{Chunking Instability Score}
We measure chunk expanding and contracting by the root-mean-square log change in the intra-chunk mean pairwise distance. For a ground-truth chunk $C$, the mean pairwise distance at recorded frame $t$ is
\begin{equation}
  D_{C}^{(t)}=\frac{2}{n_C(n_C-1)}\sum_{\substack{i<j\\i,j\in C}}\lVert x_i^{(t)}-x_j^{(t)}\rVert_2,
  \label{eq:intra_chunk_distance}
\end{equation}
where $n_C$ is the number of variables in $C$, and $x_i^{(t)}$ is the position of variable $i$. We aggregate the changes over all chunks and consecutive recorded frames as
\begin{equation}
  B_{\mathrm{RMS}}=\sqrt{\frac{1}{G(T-1)}\sum_{g=1}^{G}\sum_{t=2}^{T}\left(\log \frac{D_{g}^{(t)}}{D_{g}^{(t-1)}}\right)^2},
\end{equation}
where $G$ is the number of ground-truth chunks and $T$ is the number of recorded coordinate frames. A score of $B_{\mathrm{RMS}}=0$ indicates constant mean pairwise distances across recorded frames; lower values therefore indicate less frame-to-frame breathing.

\section{Experiment}

We evaluate Sheaf, Decentralized, Symmetrical, and Standard SyncMap on the 18 probabilistic CGCP graphs with two-state and dynamic memory. Each run uses an 80,000-state trajectory and a three-dimensional embedding; removing the initial singleton state leaves 79,999 updates. For each graph and memory setting, seeds 0--4 generate five trajectories. All methods share the trajectory for each seed, while each method's initialization is fixed across seeds; we report the mean $\pm$ sample standard deviation. Baseline hyperparameters follow \cite{Li2025}. Sheaf SyncMap uses $\lambda=1$, $b=10$, and $S=200$, fixed across graphs and seeds without graph-specific tuning. Coordinates are recorded every 100 updates, producing 799 frames. NMI uses DBSCAN with \texttt{min\_samples}=2. We compute $B_{\mathrm{RMS}}$ from the ground-truth chunks over all recorded frames. In the adaptation experiments, the final embedding coordinates and model state from each graph are used to initialize the next graph in the sequence. Because this work focuses on how sheaf regularization stabilizes chunking dynamics, we restrict comparisons to methods within the SyncMap family.

\section{Results and Discussion}
We first compare chunking performance on independent CGCP graphs under two state-memory settings and then evaluate sequential adaptation when the embedding coordinates and model state are transferred between graphs. We interpret NMI together with the chunking instability score $B_{\mathrm{RMS}}$ to distinguish accurate chunk formation from embeddings that are stable but poorly aligned with the ground-truth chunks.

\subsection{Independent CGCP Chunking Performance}
\Cref{tab:nmi_state_2,tab:nmi_state_dynamic} report NMI on the independent CGCP graphs with 2-state and dynamic state memory, respectively. With 2-state memory, Sheaf SyncMap obtains the highest mean NMI of 0.9663. Under dynamic state memory, the separation between Sheaf SyncMap and the baselines is larger. Sheaf SyncMap achieves the highest mean NMI of 0.9793, compared with 0.7608 for Decentralized SyncMap, and obtains the highest score on 17 of the 18 graphs. The only graph on which Sheaf SyncMap does not obtain the highest score is \texttt{probabilistic3\_200}, where Decentralized SyncMap reaches 0.9382 and Sheaf SyncMap reaches 0.8999. These results show that Sheaf SyncMap maintains high NMI across the two evaluated memory settings, with its largest advantage under dynamic memory and on graphs containing many small chunks.

\begin{table}[!h]
  \centering
  \resizebox{1.0\textwidth}{!}{
      \input{standalone/nmi_state_2.tex}
  }
  \caption{NMI scores for the independent CGCP graphs with 2-state memory. Values are the mean $\pm$ sample standard deviation over five seeds. Bold values mark the highest mean NMI for each graph, including ties. The final two rows report the mean score across all 18 graphs and the number of graphs on which each method achieves the highest mean. Higher values indicate better chunking performance.}
  \label{tab:nmi_state_2}
\end{table}

\begin{table}[!h]
  \centering
  \resizebox{1.0\textwidth}{!}{
      \input{standalone/nmi_state_dynamic.tex}
  }
  \caption{NMI scores for the independent CGCP graphs with dynamic state memory. Values are the mean $\pm$ sample standard deviation over five seeds. Bold values mark the highest mean NMI for each graph, including ties. The final two rows report the mean score across all 18 graphs and the number of graphs on which each method achieves the highest mean. Higher values indicate better chunking performance.}
  \label{tab:nmi_state_dynamic}
\end{table}

\subsection{Sequential Adaptation Experiments}
\begin{figure}[!h]
  \centering
  \includegraphics[width=0.9\textwidth]{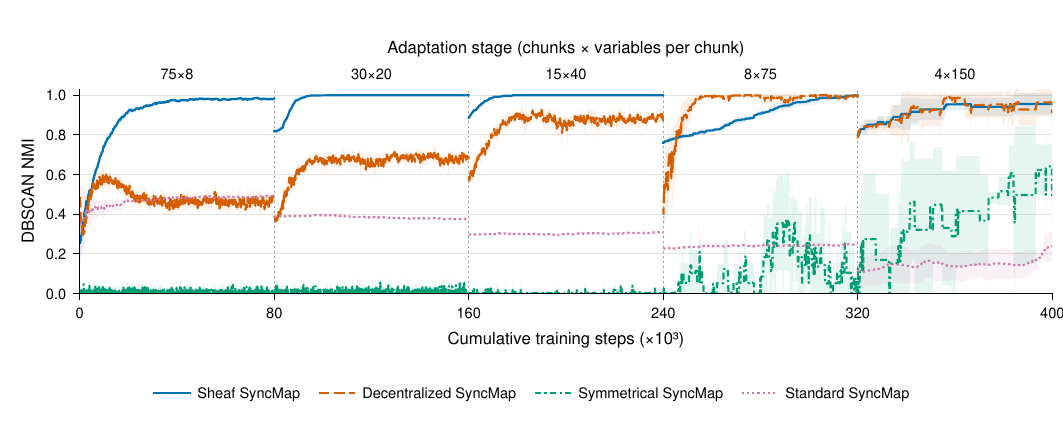}
  \caption{NMI during sequential adaptation across five CGCP graphs; each stage label gives the number of chunks $\times$ the number of variables per chunk. Lines show the mean over four seeds, shaded regions show the sample standard deviation, and vertical dotted lines mark graph transitions. Each stage is initialized with the final embedding coordinates and model state from the preceding graph. Higher values indicate better chunking performance.}
  \label{fig:adaptation_nmi}
\end{figure}

To evaluate sequential adaptation, we train each method on random walks from five probabilistic CGCP graphs in succession. The final embedding coordinates and model state from one graph initialize the next stage, so each stage transition changes the graph without resetting the learned representation. In \cref{fig:adaptation_nmi}, Sheaf SyncMap recovers high NMI after every transition: it approaches 1.0 in the first four stages and remains near 0.95 on the final \texttt{probabilistic4\_150} graph. Its performance exceeds that of Decentralized SyncMap during the first three stages and becomes comparable during the final two stages. This recovery supports the interpretation that sheaf regularization does not permanently lock the embedding to the structure learned from the preceding graph.

\begin{figure}[!h]
  \centering
  \includegraphics[width=0.9\textwidth]{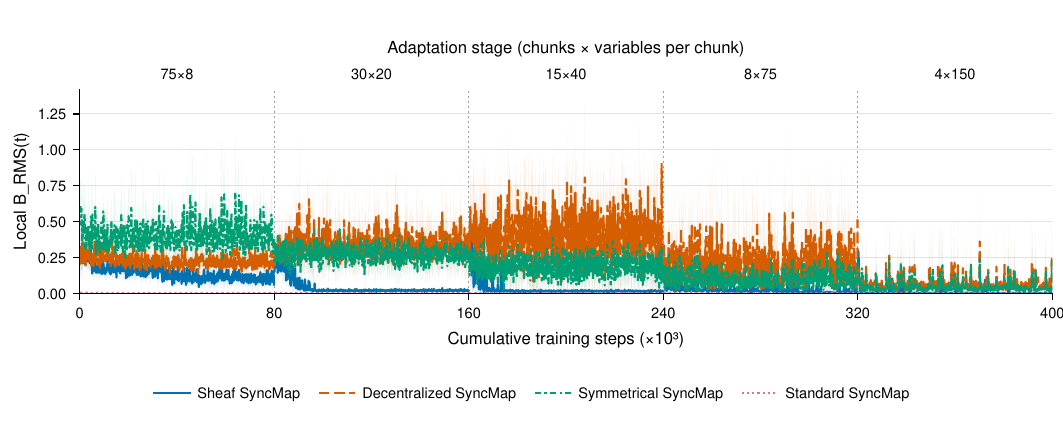}
  \caption{Local chunking instability score $B_{\mathrm{RMS}}(t)$ during sequential adaptation across five CGCP graphs; each stage label gives the number of chunks $\times$ the number of variables per chunk. Lines show the mean over four seeds, shaded regions show the sample standard deviation, and vertical dotted lines mark graph transitions. Each stage is initialized with the final embedding coordinates and model state from the preceding graph. Lower values indicate less frame-to-frame breathing.}
  \label{fig:adaptation_b_rms}
\end{figure}

\Cref{fig:adaptation_b_rms} also shows that Sheaf SyncMap generally produces less frame-to-frame breathing than Decentralized SyncMap throughout the sequence. The high NMI and reduced breathing relative to other SyncMap variants indicate that sheaf regularization stabilizes chunking dynamics without preventing adaptation to a new graph.

\subsection{Sheaf Behavior Analysis}
\begin{figure}[!h]
  \centering
  \includegraphics[width=0.9\textwidth]{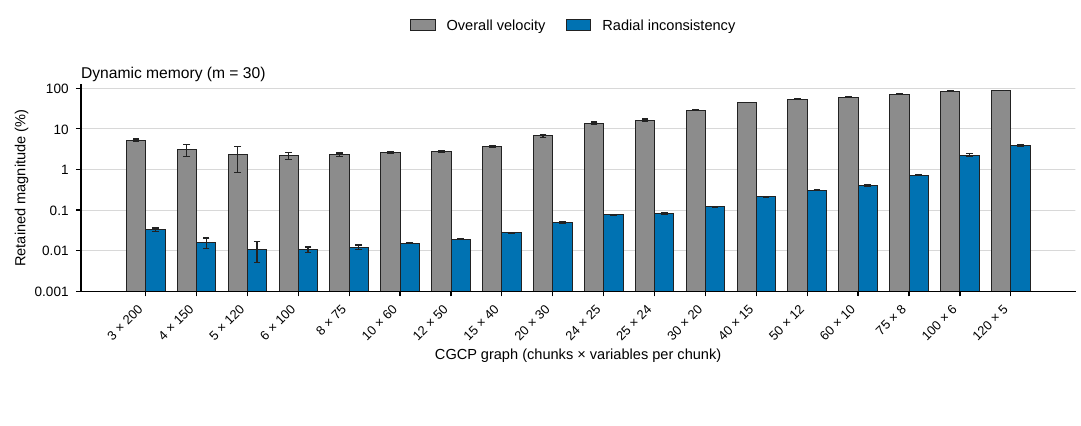}
  \caption{Velocity magnitude retained after sheaf regularization under dynamic state memory. Gray bars show the retained overall velocity magnitude, whereas blue bars show the retained radial-inconsistency magnitude. The substantially smaller radial values indicate that sheaf regularization suppresses radial inconsistency more strongly than would uniform velocity rescaling.}
  \label{fig:sheaf_velocity_retained_state_dynamic}
\end{figure}

We report the percentages of the overall velocity magnitude and radial-inconsistency magnitude retained after sheaf regularization, which are given by
\begin{equation}
  100\frac{\lVert\widetilde v_t\rVert_F}{\lVert v_t\rVert_F},
\qquad 100\sqrt{\frac{E_t(\widetilde v_t)}{E_t(v_t)}}.
  \label{eq:velocity_retained}
\end{equation}
where $E_t(v_t)=\sum_{e\in E}\omega_e s_e(v_t)^2$ is the weighted squared radial-inconsistency measure, and $\widetilde v_t$ is the sheaf-regularized velocity. \Cref{fig:sheaf_velocity_retained_state_dynamic} shows that the retained radial-inconsistency magnitude is lower than the retained overall velocity magnitude by a factor of approximately $10$--$100$, indicating that sheaf regularization suppresses radial inconsistency more strongly than would uniform velocity rescaling.

\section{Conclusion}
We proposed Sheaf SyncMap, which incorporates a radial sheaf structure to regularize pairwise forces and stabilize chunking dynamics. Experimental results on independent CGCP graphs show that Sheaf SyncMap achieves higher NMI than other SyncMap variants, particularly for graphs with many small chunks. Sequential adaptation experiments demonstrate that Sheaf SyncMap maintains high NMI while reducing frame-to-frame breathing, indicating that sheaf regularization stabilizes chunking dynamics without preventing adaptation to new graphs. Future work will explore the application of Sheaf SyncMap to real-world temporal data and investigate the theoretical properties of sheaf regularization in self-organizing systems.

\section{Acknowledgment}
This work was supported by JST SPRING (Grant No. JPMJSP2136), the JSPS Grant-in-Aid for Challenging Exploratory Research (Grant No. JP25K22833), and the JSPS Grant-in-Aid for Transformative Research Areas (A) (Grant No. JP22H05194). The authors thank their supervisor and colleagues for their valuable feedback and support.

\bibliographystyle{ACM-Reference-Format}
\bibliography{sheaf_syncmap}

\appendix
\section{Experimental Details}
We evaluate computational cost on the 18 CGCP graphs with $N=600$ variables and three-dimensional embeddings, using seeds 0--4 under both two-state and dynamic memory. Each run processes 79,999 updates. The sheaf edge set contains all $N(N-1)/2=179{,}700$ unordered pairs. We apply the radial sheaf Laplacian through edge-wise operations without explicitly constructing the coboundary matrix $B$. The implicit correction uses Jacobi-preconditioned conjugate gradient (CG), initialized with the proposed velocity $p$. Each solve is limited to 50 iterations, with the stopping criterion $\lVert r_t\rVert_2 \leq 10^{-8}+10^{-5}\lVert p\rVert_2$. If this criterion is not satisfied within the iteration budget, we retain the final iterate; the implemented update is therefore an approximate solution of \cref{eq:radial_sheaf_solution}. For $K$ CG iterations and embedding dimension $d$, the regularization requires $O((K+1)|E|d)$ time per update and $O(|E|d+Nd)$ working memory, excluding the underlying SyncMap state and recorded trajectories. With all-pairs edges, both costs grow quadratically with $N$ for fixed $K$ and $d$. 

\subsection{Ablation Study}

We conduct an ablation study to evaluate the contributions of the radial sheaf structure and directional history repulsion to the performance of Sheaf SyncMap. We compare the following four configurations:
\begin{itemize}
  \item Sheaf SyncMap (Decentralized SyncMap with both the radial sheaf structure and directional history repulsion)
  \item Decentralized SyncMap with the radial sheaf structure only
  \item Decentralized SyncMap with directional history repulsion only
  \item Decentralized SyncMap without either modification (baseline)
\end{itemize}

\begin{figure}[!h]
  \centering
  \includegraphics[width=0.9\textwidth]{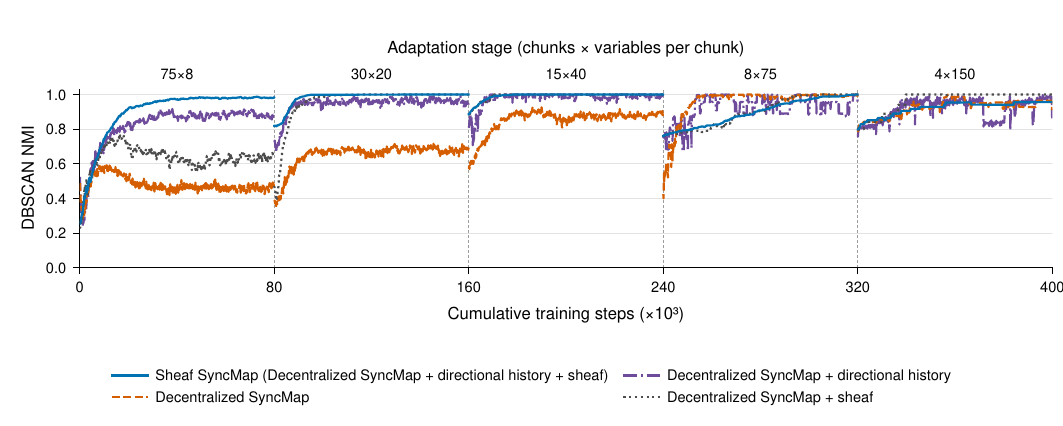}
  \caption{Ablation study of NMI scores in the adaptation experiments across five CGCP graphs.}
  \label{fig:ablation_nmi}
\end{figure}

\begin{figure}[!h]
  \centering
  \includegraphics[width=0.9\textwidth]{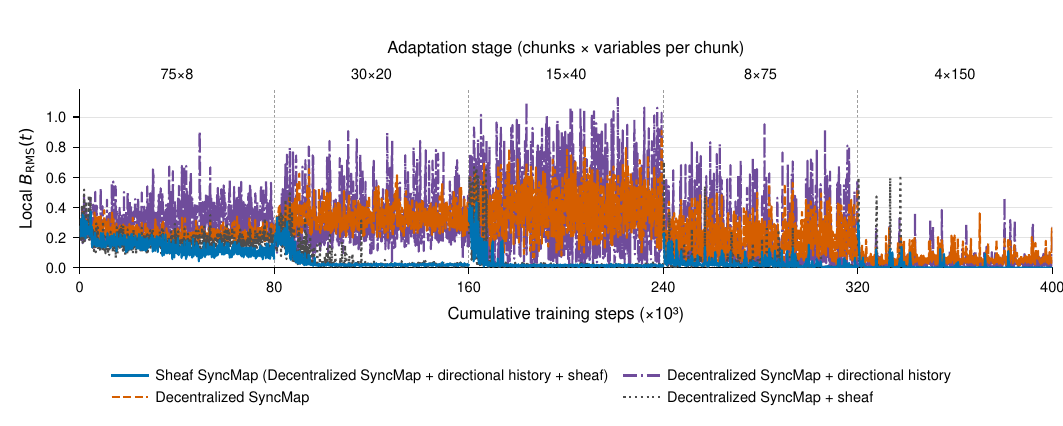}
  \caption{Ablation study of the local chunking instability score $B_{\mathrm{RMS}}(t)$ in the adaptation experiments across five CGCP graphs.}
  \label{fig:ablation_b_rms}
\end{figure}

The results of the ablation study are shown in \cref{fig:ablation_nmi,fig:ablation_b_rms}. \Cref{fig:ablation_nmi} shows that when the number of chunks is significantly larger than the number of variables per chunk, as in the $75\times8$ scenario, there is a clear performance gap among the four configurations. Their NMI scores are ordered as follows: Sheaf SyncMap $>$ Decentralized SyncMap with directional history repulsion only $>$ Decentralized SyncMap with the radial sheaf structure only $>$ Decentralized SyncMap without either modification. This result indicates that both the radial sheaf structure and directional history repulsion contribute to the performance of Sheaf SyncMap, with directional history repulsion making the larger contribution. \Cref{fig:ablation_b_rms} further shows that the radial sheaf structure contributes more strongly to the stability of the chunking dynamics across all scenarios.

\section{K-distance Heuristic for DBSCAN Clustering}
To evaluate chunking without using ground-truth labels to select the DBSCAN neighborhood radius, we apply a fixed k-distance heuristic independently to each recorded embedding frame. We set $k=3$ for radius selection and retain \texttt{min\_samples}=2 for DBSCAN. For each variable $i$ at frame $t$, we compute
\begin{equation}
  \widetilde{d}_i^{(t)}=\|x_i^{(t)}-x_{i_3}^{(t)}\|_2,
\end{equation}
where $x_{i_3}^{(t)}$ denotes the coordinates of the third-nearest variable to $i$, excluding $i$ itself. We sort these distances as $\widetilde{d}_{(1)}^{(t)}\leq\cdots\leq\widetilde{d}_{(N)}^{(t)}$, where $N$ is the number of variables. The rank and distance axes are normalized to $[0,1]$:
\begin{equation}
  u_j=\frac{j-1}{N-1},\qquad
  v_j^{(t)}=\frac{\widetilde{d}_{(j)}^{(t)}-\widetilde{d}_{(1)}^{(t)}}{\widetilde{d}_{(N)}^{(t)}-\widetilde{d}_{(1)}^{(t)}}.
\end{equation}
We then select the radius using the largest signed vertical gap below the diagonal $v=u$:
\begin{equation}
  g_j^{(t)}=u_j-v_j^{(t)},\qquad
  j_t^\star=\arg\max_j g_j^{(t)},\qquad
  \epsilon_t=\widetilde{d}_{(j_t^\star)}^{(t)}.
\end{equation}
The resulting radius is shared by all variables within frame $t$. No coordinate rescaling or curve smoothing is applied, and ground-truth labels are used only to compute NMI after clustering.

Unlike the oracle evaluation, this heuristic selects a radius independently for each frame without restricting it to that grid. The results are shown in \cref{tab:heuristic_nmi_state_2,tab:heuristic_nmi_state_dynamic}. With the k-distance heuristic, Sheaf SyncMap achieves the highest mean NMI on 9 of 18 graphs with two-state memory and on 17 of 18 graphs with dynamic memory. Under dynamic memory, its mean NMI across graphs is 0.9467, compared with 0.5187 for Decentralized SyncMap. Under two-state memory, however, Symmetrical SyncMap achieves a higher mean NMI across graphs than Sheaf SyncMap (0.9205 versus 0.9077). These results support the advantage of Sheaf SyncMap under dynamic memory with label-free radius selection.

\begin{table}[!h]
  \centering
  \resizebox{0.6\textwidth}{!}{
      \input{standalone/nmi_state_2_k_distance.tex}
  }
  \caption{NMI with two-state memory using per-frame k-distance radius selection ($k=3$, \texttt{min\_samples}=2).}
  \label{tab:heuristic_nmi_state_2}
\end{table}
\begin{table}[!h]
  \centering
  \resizebox{0.6\textwidth}{!}{
      \input{standalone/nmi_state_dynamic_k_distance.tex}
  }
  \caption{NMI with dynamic memory using per-frame k-distance radius selection ($k=3$, \texttt{min\_samples}=2).}
  \label{tab:heuristic_nmi_state_dynamic}
\end{table}
\end{document}

%% file: standalone/nmi_state_2.tex
\begin{minipage}{15.8cm}
\centering
\sffamily
\small
\setlength{\tabcolsep}{5.5pt}
\renewcommand{\arraystretch}{1.18}

\begin{tabular}{lcccc}
\toprule
Graph & \shortstack{Sheaf\\SyncMap} & \shortstack{Decentralized\\SyncMap} & \shortstack{Symmetrical\\SyncMap} & \shortstack{Standard\\SyncMap} \\
& \shortstack{(state memory:\\2)} & \shortstack{(state memory:\\2)} & \shortstack{(state memory:\\2)} & \shortstack{(state memory:\\2)} \\
\midrule
probabilistic120\_5 & \best{\meanstd{0.9332}{0.0132}} & \meanstd{0.8670}{0.0247} & \meanstd{0.0825}{0.1177} & \meanstd{0.5661}{0.0030} \\
probabilistic100\_6 & \best{\meanstd{0.9693}{0.0061}} & \meanstd{0.9207}{0.0039} & \meanstd{0.0673}{0.0734} & \meanstd{0.5389}{0.0019} \\
probabilistic75\_8 & \best{\meanstd{0.9980}{0.0013}} & \meanstd{0.9643}{0.0026} & \meanstd{0.3558}{0.0819} & \meanstd{0.5007}{0.0018} \\
probabilistic60\_10 & \best{\meanstd{1.0000}{0.0000}} & \meanstd{0.9861}{0.0031} & \meanstd{0.5795}{0.0552} & \meanstd{0.4751}{0.0018} \\
probabilistic50\_12 & \best{\meanstd{1.0000}{0.0000}} & \meanstd{0.9992}{0.0010} & \meanstd{0.6748}{0.0644} & \meanstd{0.4530}{0.0008} \\
probabilistic40\_15 & \best{\meanstd{1.0000}{0.0000}} & \best{\meanstd{1.0000}{0.0000}} & \meanstd{0.6902}{0.0614} & \meanstd{0.4245}{0.0026} \\
probabilistic30\_20 & \best{\meanstd{1.0000}{0.0000}} & \best{\meanstd{1.0000}{0.0000}} & \meanstd{0.7827}{0.0447} & \meanstd{0.3891}{0.0009} \\
probabilistic25\_24 & \best{\meanstd{1.0000}{0.0000}} & \best{\meanstd{1.0000}{0.0000}} & \meanstd{0.8525}{0.0559} & \meanstd{0.3734}{0.0018} \\
probabilistic24\_25 & \best{\meanstd{1.0000}{0.0000}} & \best{\meanstd{1.0000}{0.0000}} & \meanstd{0.8120}{0.0524} & \meanstd{0.3736}{0.0025} \\
probabilistic20\_30 & \best{\meanstd{1.0000}{0.0000}} & \best{\meanstd{1.0000}{0.0000}} & \meanstd{0.8628}{0.0683} & \meanstd{0.3430}{0.0027} \\
probabilistic15\_40 & \best{\meanstd{1.0000}{0.0000}} & \best{\meanstd{1.0000}{0.0000}} & \meanstd{0.8985}{0.0318} & \meanstd{0.3114}{0.0014} \\
probabilistic12\_50 & \best{\meanstd{1.0000}{0.0000}} & \best{\meanstd{1.0000}{0.0000}} & \best{\meanstd{1.0000}{0.0000}} & \meanstd{0.2821}{0.0010} \\
probabilistic10\_60 & \meanstd{0.9992}{0.0014} & \best{\meanstd{1.0000}{0.0000}} & \meanstd{0.9864}{0.0162} & \meanstd{0.2584}{0.0013} \\
probabilistic8\_75 & \meanstd{0.9931}{0.0154} & \best{\meanstd{1.0000}{0.0000}} & \meanstd{0.9683}{0.0189} & \meanstd{0.2344}{0.0012} \\
probabilistic6\_100 & \meanstd{0.9708}{0.0363} & \meanstd{0.9739}{0.0348} & \best{\meanstd{0.9845}{0.0169}} & \meanstd{0.1977}{0.0010} \\
probabilistic5\_120 & \meanstd{0.8555}{0.1060} & \best{\meanstd{0.9113}{0.0994}} & \meanstd{0.7772}{0.1691} & \meanstd{0.1693}{0.0012} \\
probabilistic4\_150 & \meanstd{0.8949}{0.0701} & \best{\meanstd{0.9148}{0.0692}} & \meanstd{0.4980}{0.4909} & \meanstd{0.1408}{0.0010} \\
probabilistic3\_200 & \meanstd{0.7798}{0.1032} & \best{\meanstd{0.8238}{0.1268}} & \meanstd{0.5060}{0.4759} & \meanstd{0.1090}{0.0007} \\
\midrule
Mean Score & \best{0.9663} & 0.9645 & 0.6877 & 0.3411 \\
Best count & 12 & 12 & 2 & 0 \\
\bottomrule
\end{tabular}
\end{minipage}

%% file: standalone/nmi_state_dynamic.tex
\begin{minipage}{15.8cm}
\centering
\sffamily
\small
\setlength{\tabcolsep}{5.5pt}
\renewcommand{\arraystretch}{1.18}

\begin{tabular}{lcccc}
\toprule
Graph & \shortstack{Sheaf\\SyncMap} & \shortstack{Decentralized\\SyncMap} & \shortstack{Symmetrical\\SyncMap} & \shortstack{Standard\\SyncMap} \\
& \shortstack{(state memory:\\dynamic)} & \shortstack{(state memory:\\dynamic)} & \shortstack{(state memory:\\dynamic)} & \shortstack{(state memory:\\dynamic)} \\
\midrule
probabilistic120\_5 & \best{\meanstd{0.8656}{0.0070}} & \meanstd{0.4376}{0.0086} & \meanstd{0.0051}{0.0049} & \meanstd{0.5649}{0.0058} \\
probabilistic100\_6 & \best{\meanstd{0.8990}{0.0097}} & \meanstd{0.4297}{0.0147} & \meanstd{0.0049}{0.0040} & \meanstd{0.5304}{0.0065} \\
probabilistic75\_8 & \best{\meanstd{0.9816}{0.0089}} & \meanstd{0.4708}{0.0120} & \meanstd{0.0047}{0.0050} & \meanstd{0.4911}{0.0048} \\
probabilistic60\_10 & \best{\meanstd{0.9974}{0.0026}} & \meanstd{0.5250}{0.0243} & \meanstd{0.0068}{0.0054} & \meanstd{0.4577}{0.0056} \\
probabilistic50\_12 & \best{\meanstd{0.9988}{0.0015}} & \meanstd{0.5604}{0.0131} & \meanstd{0.0082}{0.0048} & \meanstd{0.4432}{0.0021} \\
probabilistic40\_15 & \best{\meanstd{1.0000}{0.0001}} & \meanstd{0.6218}{0.0198} & \meanstd{0.0063}{0.0060} & \meanstd{0.4129}{0.0045} \\
probabilistic30\_20 & \best{\meanstd{0.9999}{0.0001}} & \meanstd{0.6930}{0.0179} & \meanstd{0.0098}{0.0044} & \meanstd{0.3770}{0.0038} \\
probabilistic25\_24 & \best{\meanstd{1.0000}{0.0000}} & \meanstd{0.7271}{0.0161} & \meanstd{0.0050}{0.0011} & \meanstd{0.3656}{0.0074} \\
probabilistic24\_25 & \best{\meanstd{1.0000}{0.0000}} & \meanstd{0.7354}{0.0306} & \meanstd{0.0031}{0.0030} & \meanstd{0.3626}{0.0030} \\
probabilistic20\_30 & \best{\meanstd{1.0000}{0.0000}} & \meanstd{0.7893}{0.0251} & \meanstd{0.0019}{0.0023} & \meanstd{0.3394}{0.0048} \\
probabilistic15\_40 & \best{\meanstd{1.0000}{0.0000}} & \meanstd{0.9125}{0.0077} & \meanstd{0.0023}{0.0018} & \meanstd{0.3033}{0.0061} \\
probabilistic12\_50 & \best{\meanstd{1.0000}{0.0000}} & \meanstd{0.9689}{0.0098} & \meanstd{0.0027}{0.0018} & \meanstd{0.2777}{0.0058} \\
probabilistic10\_60 & \best{\meanstd{1.0000}{0.0000}} & \meanstd{0.9895}{0.0062} & \meanstd{0.0449}{0.0566} & \meanstd{0.2577}{0.0052} \\
probabilistic8\_75 & \best{\meanstd{1.0000}{0.0000}} & \meanstd{0.9980}{0.0021} & \meanstd{0.0788}{0.1731} & \meanstd{0.2460}{0.0065} \\
probabilistic6\_100 & \best{\meanstd{1.0000}{0.0000}} & \meanstd{0.9844}{0.0301} & \meanstd{0.2676}{0.2597} & \meanstd{0.2268}{0.0086} \\
probabilistic5\_120 & \best{\meanstd{1.0000}{0.0000}} & \meanstd{0.9422}{0.1125} & \meanstd{0.3740}{0.3461} & \meanstd{0.2316}{0.0405} \\
probabilistic4\_150 & \best{\meanstd{0.9845}{0.0215}} & \meanstd{0.9707}{0.0635} & \meanstd{0.6596}{0.1286} & \meanstd{0.2325}{0.0611} \\
probabilistic3\_200 & \meanstd{0.8999}{0.0911} & \best{\meanstd{0.9382}{0.1158}} & \meanstd{0.5366}{0.3041} & \meanstd{0.2886}{0.0976} \\
\midrule
Mean Score & \best{0.9793} & 0.7608 & 0.1124 & 0.3560 \\
Best count & 17 & 1 & 0 & 0 \\
\bottomrule
\end{tabular}
\end{minipage}

%% file: standalone/nmi_state_2_k_distance.tex
\begin{minipage}{15.8cm}
\centering
\sffamily
\small
\setlength{\tabcolsep}{5.5pt}
\renewcommand{\arraystretch}{1.18}

\par\smallskip
\begin{tabular}{lcccc}
\toprule
Graph & \shortstack{Sheaf\\SyncMap} & \shortstack{Decentralized\\SyncMap} & \shortstack{Symmetrical\\SyncMap} & \shortstack{Standard\\SyncMap} \\
& \shortstack{(state memory:\\2)} & \shortstack{(state memory:\\2)} & \shortstack{(state memory:\\2)} & \shortstack{(state memory:\\2)} \\
\midrule
probabilistic120\_5 & \meanstd{0.5600}{0.2214} & \meanstd{0.0205}{0.0060} & \best{\meanstd{0.9483}{0.0052}} & \meanstd{0.0682}{0.0080} \\
probabilistic100\_6 & \meanstd{0.9457}{0.0150} & \meanstd{0.7681}{0.1626} & \best{\meanstd{0.9544}{0.0061}} & \meanstd{0.0589}{0.0131} \\
probabilistic75\_8 & \best{\meanstd{0.9585}{0.0130}} & \meanstd{0.8833}{0.0146} & \meanstd{0.9454}{0.0082} & \meanstd{0.0564}{0.0006} \\
probabilistic60\_10 & \meanstd{0.8975}{0.0148} & \meanstd{0.9060}{0.0069} & \best{\meanstd{0.9382}{0.0089}} & \meanstd{0.0599}{0.0058} \\
probabilistic50\_12 & \meanstd{0.8886}{0.0021} & \meanstd{0.9200}{0.0031} & \best{\meanstd{0.9342}{0.0064}} & \meanstd{0.0657}{0.0067} \\
probabilistic40\_15 & \meanstd{0.9139}{0.0198} & \best{\meanstd{0.9340}{0.0044}} & \meanstd{0.9326}{0.0072} & \meanstd{0.0586}{0.0055} \\
probabilistic30\_20 & \best{\meanstd{0.9435}{0.0097}} & \meanstd{0.9265}{0.0095} & \meanstd{0.9177}{0.0174} & \meanstd{0.0476}{0.0045} \\
probabilistic25\_24 & \best{\meanstd{0.9717}{0.0112}} & \meanstd{0.9354}{0.0069} & \meanstd{0.9120}{0.0125} & \meanstd{0.0501}{0.0048} \\
probabilistic24\_25 & \best{\meanstd{0.9770}{0.0094}} & \meanstd{0.9387}{0.0139} & \meanstd{0.9208}{0.0121} & \meanstd{0.0543}{0.0058} \\
probabilistic20\_30 & \best{\meanstd{0.9873}{0.0113}} & \meanstd{0.9577}{0.0133} & \meanstd{0.8990}{0.0087} & \meanstd{0.0444}{0.0000} \\
probabilistic15\_40 & \best{\meanstd{0.9820}{0.0111}} & \meanstd{0.9780}{0.0141} & \meanstd{0.8865}{0.0154} & \meanstd{0.0414}{0.0048} \\
probabilistic12\_50 & \meanstd{0.9827}{0.0079} & \best{\meanstd{0.9837}{0.0033}} & \meanstd{0.8942}{0.0151} & \meanstd{0.0379}{0.0049} \\
probabilistic10\_60 & \best{\meanstd{0.9940}{0.0068}} & \meanstd{0.9776}{0.0076} & \meanstd{0.9214}{0.0509} & \meanstd{0.0367}{0.0021} \\
probabilistic8\_75 & \best{\meanstd{0.9895}{0.0155}} & \meanstd{0.9762}{0.0188} & \meanstd{0.9556}{0.0360} & \meanstd{0.0340}{0.0034} \\
probabilistic6\_100 & \best{\meanstd{0.9710}{0.0339}} & \meanstd{0.9689}{0.0316} & \meanstd{0.9542}{0.0210} & \meanstd{0.0253}{0.0044} \\
probabilistic5\_120 & \meanstd{0.8236}{0.1091} & \meanstd{0.8696}{0.1100} & \best{\meanstd{0.9379}{0.0307}} & \meanstd{0.0265}{0.0002} \\
probabilistic4\_150 & \meanstd{0.8195}{0.1633} & \best{\meanstd{0.8976}{0.0794}} & \meanstd{0.8788}{0.0636} & \meanstd{0.0275}{0.0042} \\
probabilistic3\_200 & \meanstd{0.7322}{0.0317} & \meanstd{0.8020}{0.1160} & \best{\meanstd{0.8380}{0.2032}} & \meanstd{0.0115}{0.0010} \\
\midrule
Mean Score & 0.9077 & 0.8691 & \best{0.9205} & 0.0447 \\
Best count & 9 & 3 & 6 & 0 \\
\bottomrule
\end{tabular}
\end{minipage}

%% file: standalone/nmi_state_dynamic_k_distance.tex
\begin{minipage}{15.8cm}
\centering
\sffamily
\small
\setlength{\tabcolsep}{5.5pt}
\renewcommand{\arraystretch}{1.18}

\par\smallskip
\begin{tabular}{lcccc}
\toprule
Graph & \shortstack{Sheaf\\SyncMap} & \shortstack{Decentralized\\SyncMap} & \shortstack{Symmetrical\\SyncMap} & \shortstack{Standard\\SyncMap} \\
& \shortstack{(state memory:\\dynamic)} & \shortstack{(state memory:\\dynamic)} & \shortstack{(state memory:\\dynamic)} & \shortstack{(state memory:\\dynamic)} \\
\midrule
probabilistic120\_5 & \best{\meanstd{0.7802}{0.0672}} & \meanstd{0.0740}{0.0144} & \meanstd{0.1294}{0.0180} & \meanstd{0.0675}{0.0084} \\
probabilistic100\_6 & \best{\meanstd{0.8780}{0.0099}} & \meanstd{0.0794}{0.0118} & \meanstd{0.1409}{0.0182} & \meanstd{0.0449}{0.0127} \\
probabilistic75\_8 & \best{\meanstd{0.9761}{0.0104}} & \meanstd{0.0859}{0.0079} & \meanstd{0.1817}{0.0348} & \meanstd{0.0732}{0.0290} \\
probabilistic60\_10 & \best{\meanstd{0.9933}{0.0015}} & \meanstd{0.0726}{0.0154} & \meanstd{0.1998}{0.0247} & \meanstd{0.0454}{0.0158} \\
probabilistic50\_12 & \best{\meanstd{0.9920}{0.0014}} & \meanstd{0.0925}{0.0241} & \meanstd{0.2329}{0.0705} & \meanstd{0.0479}{0.0088} \\
probabilistic40\_15 & \best{\meanstd{0.9878}{0.0023}} & \meanstd{0.1094}{0.0315} & \meanstd{0.2706}{0.0540} & \meanstd{0.0779}{0.0048} \\
probabilistic30\_20 & \best{\meanstd{0.9921}{0.0017}} & \meanstd{0.1447}{0.0563} & \meanstd{0.2953}{0.0314} & \meanstd{0.0607}{0.0334} \\
probabilistic25\_24 & \best{\meanstd{0.9866}{0.0069}} & \meanstd{0.3364}{0.1003} & \meanstd{0.3033}{0.0667} & \meanstd{0.0303}{0.0185} \\
probabilistic24\_25 & \best{\meanstd{0.9866}{0.0038}} & \meanstd{0.3395}{0.0294} & \meanstd{0.2989}{0.0618} & \meanstd{0.0567}{0.0184} \\
probabilistic20\_30 & \best{\meanstd{0.9844}{0.0089}} & \meanstd{0.6202}{0.0728} & \meanstd{0.4067}{0.0464} & \meanstd{0.0347}{0.0155} \\
probabilistic15\_40 & \best{\meanstd{0.9784}{0.0093}} & \meanstd{0.9066}{0.0052} & \meanstd{0.5017}{0.0427} & \meanstd{0.0639}{0.0208} \\
probabilistic12\_50 & \best{\meanstd{0.9721}{0.0093}} & \meanstd{0.9576}{0.0126} & \meanstd{0.5558}{0.0324} & \meanstd{0.0310}{0.0018} \\
probabilistic10\_60 & \best{\meanstd{0.9796}{0.0042}} & \meanstd{0.9618}{0.0046} & \meanstd{0.6072}{0.1101} & \meanstd{0.0212}{0.0103} \\
probabilistic8\_75 & \best{\meanstd{0.9712}{0.0068}} & \meanstd{0.9619}{0.0070} & \meanstd{0.7006}{0.0252} & \meanstd{0.0333}{0.0090} \\
probabilistic6\_100 & \best{\meanstd{0.9544}{0.0207}} & \meanstd{0.9435}{0.0260} & \meanstd{0.7311}{0.0210} & \meanstd{0.0333}{0.0262} \\
probabilistic5\_120 & \best{\meanstd{0.9489}{0.0131}} & \meanstd{0.9225}{0.1024} & \meanstd{0.7193}{0.0453} & \meanstd{0.0439}{0.0376} \\
probabilistic4\_150 & \best{\meanstd{0.9407}{0.0216}} & \meanstd{0.8833}{0.1183} & \meanstd{0.7527}{0.0277} & \meanstd{0.0287}{0.0224} \\
probabilistic3\_200 & \meanstd{0.7389}{0.0673} & \best{\meanstd{0.8454}{0.2786}} & \meanstd{0.7834}{0.0307} & \meanstd{0.0875}{0.1438} \\
\midrule
Mean Score & \best{0.9467} & 0.5187 & 0.4340 & 0.0490 \\
Best count & 17 & 1 & 0 & 0 \\
\bottomrule
\end{tabular}
\end{minipage}